\documentclass[runningheads]{llncs}
\usepackage{graphicx}
\usepackage{amsmath,amssymb} 
\usepackage{color}
\usepackage{booktabs}
\usepackage{algorithm}
\usepackage{algorithmicx}
\usepackage{algpseudocode}
\usepackage{epsfig}
\usepackage[width=122mm,left=12mm,paperwidth=146mm,height=193mm,top=12mm,paperheight=217mm]{geometry}

\usepackage{multirow}
\usepackage{array}

\newcommand{\etal}{\textit{et al}.}
\newcommand{\ie}{\textit{i}.\textit{e}.}
\newcommand{\eg}{\textit{e}.\textit{g}.}

\newcolumntype{C}{ >{\centering\arraybackslash} m{4cm} }
\newcolumntype{D}{ >{\centering\arraybackslash} m{1cm} }

\begin{document}
\title{Person Re-identification with\\ Deep Similarity-Guided Graph Neural Network} 

\titlerunning{Person Re-ID with Deep Similarity-Guided Graph Neural Network}
%
\author{Yantao Shen\inst{1}\and
Hongsheng Li\inst{1}\thanks{Hongsheng Li is the corresponding author.} \and 
Shuai Yi\inst{2} \and \\
Dapeng Chen\inst{1} \and
Xiaogang Wang\inst{1}}
%
\authorrunning{Y. Shen, H. Li, S. Yi, D. Chen and X. Wang}
%
\institute{CUHK-SenseTime Joint Lab, The Chinese University of Hong Kong
\email{\{ytshen,hsli,dpchen,xgwang\}@ee.cuhk.edu.hk}\\\and
SenseTime Research\\
\email{yishuai@sensetime.com}}
\maketitle              
\begin{abstract}
The person re-identification task requires to robustly estimate visual similarities between person images. However, existing person re-identification models mostly estimate the similarities of different image pairs of probe and gallery images independently while ignores the relationship information between different probe-gallery pairs. As a result, the similarity estimation of some hard samples might not be accurate. In this paper, we propose a novel deep learning framework, named Similarity-Guided Graph Neural Network (SGGNN) to overcome such limitations. Given a probe image and several gallery images, SGGNN creates a graph to represent the pairwise relationships between probe-gallery pairs (nodes) and utilizes such relationships to update the probe-gallery relation features in an end-to-end manner. Accurate similarity estimation can be achieved by using such updated probe-gallery relation features for prediction. The input features for nodes on the graph are the relation features of different probe-gallery image pairs. The probe-gallery relation feature updating is then performed by the messages passing in SGGNN, which takes other nodes' information into account for similarity estimation. Different from conventional GNN approaches, SGGNN learns the edge weights with rich labels of gallery instance pairs directly, which provides relation fusion more precise information. The effectiveness of our proposed method is validated on three public person re-identification datasets.

\end{abstract}
\begin{figure}[t]
\centering
\begin{tabular}{cc}
   & \includegraphics[scale = 0.25]{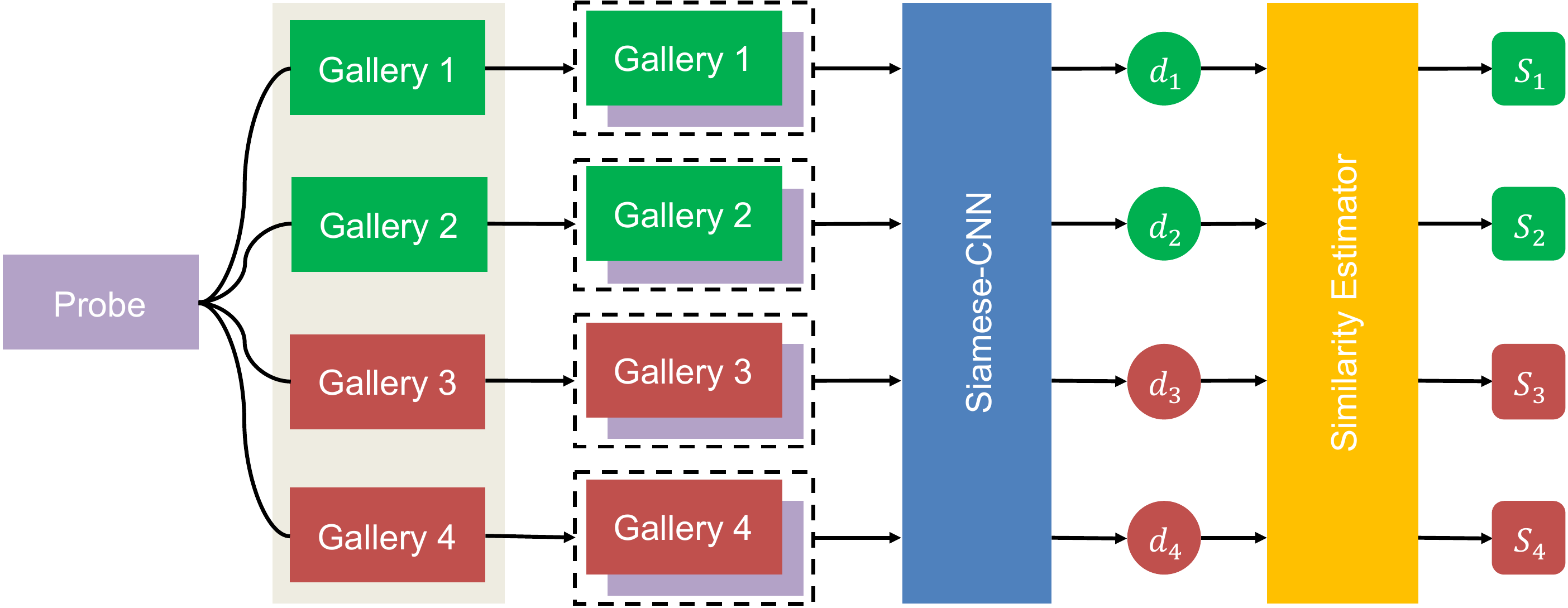}\\
   & (a) Conventional Approach.\\
   & \includegraphics[scale = 0.25]{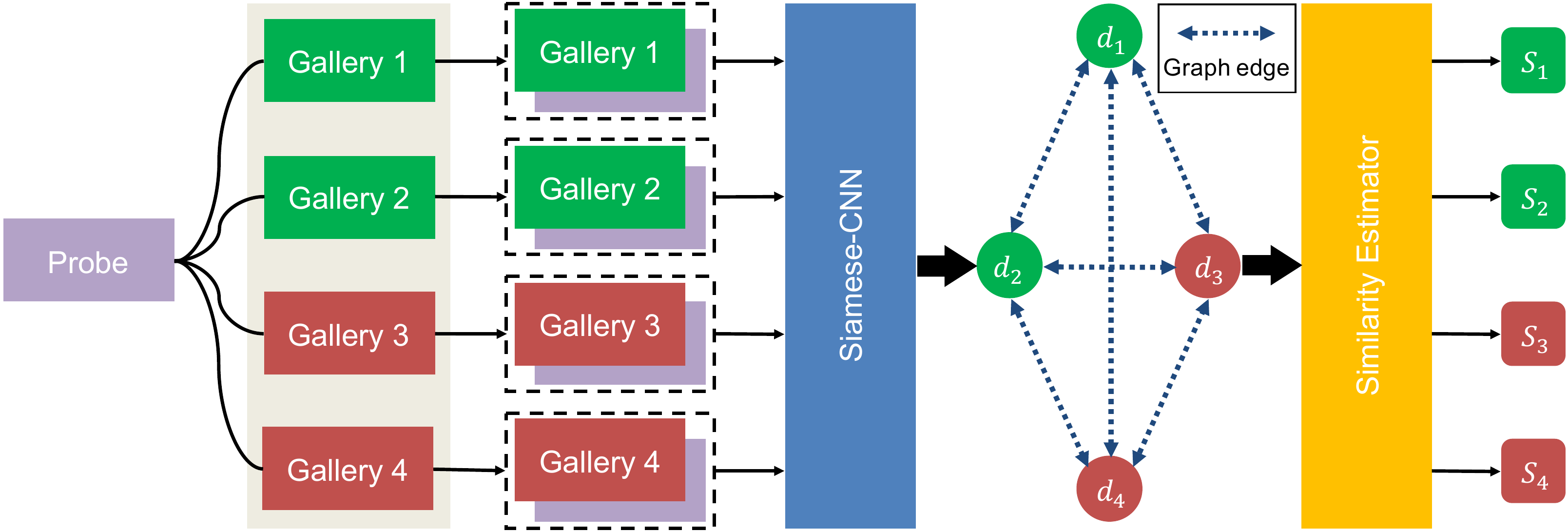}\\
   & (b) Our proposed SGGNN.
\end{tabular}
\caption{Illustration of our Proposed SGGNN method and conventional person re-identification approach. (a) The pipeline of conventional person re-identification approach, the pairwise relationships between different probe-gallery pairs are ignored. The similarity score of each probe-gallery pair $d_i$ ($i=1,2,3,4$) is estimated individually. (b) Our proposed SGGNN approach, pairwise relationships between different probe-gallery pairs are involved with deeply learned message passing on a graph for more accurate similarity estimation.}
\label{fig:main}
\end{figure}

\section{Introduction}
Person re-identification is a challenging problem, which aims at finding the person images of interest in a set of images across different cameras. It plays a significant role in the intelligent surveillance systems. 

To enhance the re-identification performance, most existing approaches attempt to learn discriminative features or design various metric distances for better measuring the similarities between person image pairs. In recent years, witness the success of deep learning based approaches for various tasks of computer vision~\cite{krizhevsky2012imagenet,he2016deep,ren2015faster,wu2016model,srivastava2014dropout,chu2017multi,Liuxihui_2017_ICCV,wu20173d,yang2017learning,li2017vip,kang2017object}, a large number of deep learning methods were proposed for person re-identification~\cite{liu2016multi,Zhou_2017_CVPR,xiao2016learning,liu2017quality}. Most of these deep learning based approaches utilized Convolutional Neural Network~(CNN) to learn robust and discriminative features. In the mean time, metric learning methods were also proposed~\cite{Bak_2017_CVPR,bkak2016person,Yu_2017_ICCV} to generate relatively small feature distances between images of same identity and large feature distances between those of different identities.

However, most of these approaches only consider the pairwise similarity while ignore the internal similarities among the images of the whole set. For instance, when we attempt to estimate the similarity score between a probe image and a gallery image, most feature learning and metric learning approaches only consider the pairwise relationship between this single probe-gallery image pair in both training and testing stages. Other relations among different pairs of images are ignored. As a result, some hard positive or hard negative pairs are difficult to obtain proper similarity scores since only limited relationship information among samples is utilized for similarity estimation.

To overcome such limitation, we need to discover the valuable internal similarities among the image set, especially for the similarities among the gallery set. One possible solution is utilizing manifold learning~\cite{bai2017scalable,loy2013person}, which considers the similarities of each pair of images in the set. It maps images into a manifold with more smooth local geometry. Beyond the manifold learning methods, re-ranking approaches~\cite{Zhong_2017_CVPR,garcia2015person,ye2016person} were also utilized for refining the ranking result by integrating similarities between top-ranked gallery images. However, both manifold learning and re-ranking approaches have two major limitations: (1) most manifold learning and re-ranking approaches are unsupervised, which could not fully exploit the provided training data label into the learning process. (2) These two kinds of approaches could not benefit feature learning since they are not involved in training process.  

Recently, \textit{Graph Neural Network}~(GNN)~\cite{bruna2013spectral,henaff2015deep,kipf2016semi,niepert2016learning} draws increasing attention due to its ability of generalizing neural networks for data with graph structures. The GNN propagates messages on a graph structure. After message traversal on the graph, node's final representations are obtained from its own as well as other node's information, and are then utilized for node classification. GNN has achieved huge success in many research fields, such as text classification~\cite{defferrard2016convolutional}, image classification~\cite{bruna2013spectral,oord2016pixel}, and human action recognition~\cite{yan2018spatial}. Compared with manifold learning and re-ranking, GNN incorporates graph computation into the neural networks learning, which makes the training end-to-end and benefits learning the feature representation.

In this paper, we propose a novel deep learning framework for person re-identification, named Similarity-Guided Graph Neural Network (SGGNN). SGGNN incorporates graph computation in both training and testing stages of deep networks for obtaining robust similarity estimations and discriminative feature representations. Given a mini-batch consisting of several probe images and gallery images, SGGNN will first learn initial visual features for each image (\eg, global average pooled features from ResNet-50~\cite{he2016deep}.) with the pairwise relation supervisions. After that,  each pair of probe-gallery images will be treated as a node on the graph, which is responsible for generating similarity score of this pair. To fully utilize pairwise relations between other pairs (nodes) of images, deeply learned messages are propagated among nodes to update and refine the pairwise relation features associated with each node. Unlike most previous GNNs' designs, in SGGNN, the weights for feature fusion are determined by similarity scores by gallery image pairs, which are directly supervised by training labels. With these similarity guided feature fusion weights, SGGNN will fully exploit the valuable label information to generate discriminative person image features and obtain robust similarity estimations for probe-gallery image pairs. 

The main contribution of this paper is two-fold. (1) We propose a novel Similarity Guided Graph Neural Network (SGGNN) for person re-identification, which could be trained end-to-end. Unlike most existing methods, which utilize inter-gallery-image relations between samples in the post-processing stage, SGGNN incorporates the inter-gallery-image relations in the training stage to enhance feature learning process. As a result, more discriminative and accurate person image feature representations could be learned. (2) Different from most Graph Neural Network (GNN) approaches, SGGNN exploits the training label supervision for learning more accurate feature fusion weights for updating the nodes' features. This similarity guided manner ensures the feature fusion weights to be more precise and conduct more reasonable feature fusion. The effectiveness of our proposed method is verified by extensive experiments on three large person re-identification datasets.

\section{Related Work}

\subsection{Person Re-identification}
Person re-identification is an active research topic, which gains increasing attention from both academia and industry in recent years. The mainstream approaches for person re-identification either try to obtain discriminative and robust feature~\cite{yi2014deep,li2014deepreid,ahmed2015improved,Su_2017_ICCV,schumann2017person,cheng2016person,Lin_2017_CVPR,Sun_2017_ICCV,shen2017learning,shen2018deep,chen2018group,chen2018video,song2017region,karaman2014leveraging} for representing person image or design a proper metric distance for measuring similarity between person images~\cite{paisitkriangkrai2015learning,bkak2016person,Bak_2017_CVPR,Liu_2017_ICCV,Yu_2017_ICCV}. For feature learning, Yi \etal~\cite{yi2014deep} introduced a Siamese-CNN for person re-identification.  Li \etal~\cite{li2014deepreid} proposed a novel filter pairing neural network, which could jointly handle feature learning, misalignment, and classification in an end-to-end manner.  Ahmed \etal~\cite{ahmed2015improved} introduced a model called Cross-Input Neighbourhood Difference CNN model, which compares image features in each patch of one input image to the other image's patch. Su \etal~\cite{Su_2017_ICCV} incorporated pose information into person re-identification. The pose estimation algorithm are utilized for part extraction. Then the original global image and the transformed part images are fed into a CNN simultaneously for prediction. Shen \etal~\cite{shen2018end} utilized kronecker-product matching for person feature maps alignment. For metric learning, Paisitkriangkrai \etal~\cite{paisitkriangkrai2015learning} introduced an approach aims at learning the weights of different metric distance functions by optimizing the relative distance among triplet samples and maximizing the averaged rank-k accuracies. Bak \etal~\cite{bkak2016person} proposed to learn metrics for 2D patches of person image. Yu \etal~\cite{Yu_2017_ICCV} introduced an  unsupervised person re-ID model, which aims at learning an asymmetric metric on cross-view person images.

Besides feature learning and metric learning, manifold learning~\cite{bai2017scalable,loy2013person} and re-rank approaches~\cite{Zhong_2017_CVPR,ye2015ranking,ye2016person,garcia2015person} are also utilized for enhancing the performance of person re-identification model, Bai \etal~\cite{bai2017scalable} introduced Supervised Smoothed Manifold, which aims to estimating the context of other pairs of person image thus the learned relationships with between samples are smooth on the manifold. Loy \etal~\cite{loy2013person} introduced manifold ranking for revealing manifold structure by plenty of gallery images. Zhong \etal~\cite{Zhong_2017_CVPR} utilized k-reciprocal encoding to refine the ranking list result by exploiting relationships between top rank gallery instances for a probe sample. Kodirov \etal~\cite{kodirov2016person} introduced graph regularised dictionary learning for person re-identification. Most of these approaches are conducted in the post-process stage and the visual features of person images could not be benefited from these post-processing approaches.

\subsection{Graph for Machine Learning}
\label{subsec:graph}
In several machine learning research areas, input data could be naturally represented as graph structure, such as natural language processing~\cite{mills2014graph,liu2018show}, human pose estimation~\cite{chu2016crf,yan2018spatial,yang2016end}, visual relationship detection~\cite{li2017scene}, and image classification~\cite{quek2011structural,pavlidis2013structural}. In~\cite{scarselli2009graph}, Scarselli \etal~divided machine learning models into two classes due to different application objectives on graph data structure, named \textit{node-focused} and \textit{graph-focused} application. For \textit{graph-focused} application, the mapping function takes the whole graph data $G$ as the input. One simple example for \textit{graph-focused} application is to classify the image~\cite{pavlidis2013structural}, where the image is represented by a region adjacency graph. For \textit{node-focused} application, the inputs of mapping function are the nodes on the graph. Each node on the graph will represent a sample in the dataset and the edge weights will be determined by the relationships between samples. After the message propagation among different nodes (samples), the mapping function will output the classification or regression results of each node. One typical example for \textit{node-focused} application is graph based image segmentation~\cite{zheng2015conditional,liu2015crf}, which takes pixels of image as nodes and try to minimize the total energy function for segmentation prediction of each pixel. Another example for \textit{node-focused} application is object detection~\cite{bianchini2005recursive}, the input nodes are features of the proposals in a input image.

\subsection{Graph Neural Network}
Scarselli \etal~\cite{scarselli2009graph} introduced Graph Neural Network (GNN), which is an extension for recursive neural networks and random walk models for graph structure data. It could be applied for both graph-focused or node-focused data without any pre or post-processing steps, which means that it can be trained end-to-end. In recent years, extending CNN to graph data structure received increased attention~\cite{bruna2013spectral,henaff2015deep,kipf2016semi,niepert2016learning,yan2018spatial,defferrard2016convolutional,liang2016semantic},  Bruna \etal~\cite{bruna2013spectral} proposed two constructions of deep convolutional networks on graphs (GCN), one is based on the spectrum of graph Laplacian, which is called spectral construction. Another is spatial construction, which extends properties of convolutional filters to general graphs. Yan \etal~\cite{yan2018spatial} exploited spatial construction GCN for human action recognition. Different from most existing GNN approaches, our proposed approach exploits the training data label supervision for generating more accurate feature fusion weights in the graph message passing.

\section{Method} 
\label{sec:method}

To evaluate the algorithms for person re-identification, the test dataset is usually divided into two parts: a probe set and a gallery set. Given an image pair of a probe and a gallery images, the person re-identification models aims at robustly determining visual similarities between probe-gallery image pairs. In the previous common settings, among a mini-batch, different image pairs of probe and gallery images are evaluated individually, \ie, the estimated similarity between a pair of images will not be influenced by other pairs. However, the similarities between different gallery images are valuable for refining similarity estimation between the probe and gallery. Our proposed approach is proposed to better utilize such information to improve feature learning and is illustrated in Figure~\ref{fig:main}. It takes a probe and several gallery images as inputs to create a graph with each node modeling a probe-gallery image pair. It outputs the similarity score of each probe-gallery image pair. Deeply learned messages will be propagated among nodes to update the relation features associated with each node for more accurate similarity score estimation in the end-to-end training process. 

In this section, the problem formulation and node features will be discussed in Section \ref{sec:pro_for}. The Similarity Guided GNN (SGGNN) and deep messages propagation for person re-identification will be presented in Section \ref{sec:SGGNN}. Finally, we will discuss the advantage of similarity guided edge weight over the conventional GNN approaches in Section \ref{sec:advan}. The implementation details will be introduced in \ref{sec:imp}

\subsection{Graph Formulation and Node Features}
\label{sec:pro_for}

In our framework, we formulate person re-identification as a \textit{node-focused} graph application introduced in Section~\ref{subsec:graph}. Given a probe image and $N$ gallery images, we construct an undirected complete graph $G(V,E)$, where $V = \{v_1, v_2, ..., v_N\}$ denotes the set of nodes. Each node represents a pair of probe-gallery images. Our goal is to estimate the similarity score for each probe-gallery image pair and therefore treat the re-identification problem as a node classification problem. Generally, the input features for any node encodes the complex relations between its corresponding probe-gallery image pair. 

In this work, we adopt a simple approach for obtaining input relation features to the graph nodes, which is shown in Figure~\ref{fig:embed}(a). Given a probe image and $N$ gallery images, each input probe-gallery image pair will be fed into a Siamese-CNN for pairwise relation feature encoding. The Siamese-CNN's structure is based on the ResNet-50~\cite{he2016deep}. To obtain the pairwise relation features, the last global average pooled features of two images from ResNet-50 are element-wise subtracted. The pairwise feature is processed by element-wise square operation and a Batch Normalization layer~\cite{ioffe2015batch}. The processed difference features $d_i$ ($i=1,2,...,N$) encode the deep visual relations between the probe and the $i$-th gallery image, and are used as the input features of the $i$-th node on the graph. Since our task is node-wise classification, \ie, estimating the similarity score of each probe-gallery pair, a naive approach would be simply feeding each node's input feature into a linear classifier to output the similarity score without considering the pairwise relationship between different nodes. For each probe-gallery image pair in the training mini-batch, a binary cross-entropy loss function could be utilized,
\begin{equation}\label{eq:loss}
   L =-\sum_{i=1}^{N}y_i \log(f(d_i))+(1-y_i)\log(1-f(d_i)) ,
\end{equation}
where $f()$ denotes a linear classifier followed by a sigmoid function. $y_i$ denotes the ground-truth label of $i$-th probe-gallery image pair, with 1 representing the probe and the $i$-th gallery images belonging to the same identity while 0 for not.

\subsection{Similarity-Guided Graph Neural Network}
\label{sec:SGGNN}
Obviously, the naive node classification model (Eq.(~\ref{eq:loss})) ignores the valuable information among different probe-gallery pairs. For exploiting such vital information, we need to establish edges $E$ on the graph $G$. In our formulation, $G$ is fully-connected and $E$ represents the set of relationships between different probe-gallery pairs, where $W_{ij}$ is a scalar edge weight. It represents the relation importance between node~$i$ and node~$j$ and can be calculated as,
\begin{equation}\label{eq:egde_weight}
   W_{ij} = 
   \begin{cases}
   \frac{\text{exp}(S(g_i,g_j))}{\sum_{j}\text{exp}(S(g_i,g_j))}, \quad i \neq j \\
   0, \quad i = j\\
   \end{cases},
\end{equation}
where $g_i$ and $g_j$ are the $i$-th and $j$-th gallery images. $S()$ is a pairwise similarity estimation function, that estimates the similarity score between $g_i$ and $g_j$ and can be modeled in the same way as the naive node (probe-gallery image pair) classification model discussed above. Note that in SGGNN, the similarity score $S(g_i,g_j)$ of gallery-gallery pair is also learned in a supervised way with person identity labels. The purpose of setting $W_{ii}$ to 0 is to avoid self-enhancing. 
\begin{figure}[t]
\centering
\begin{tabular}{m{0cm}@{\hspace{3mm}}m{6cm}m{7cm}}
   & \includegraphics[scale = 0.4]{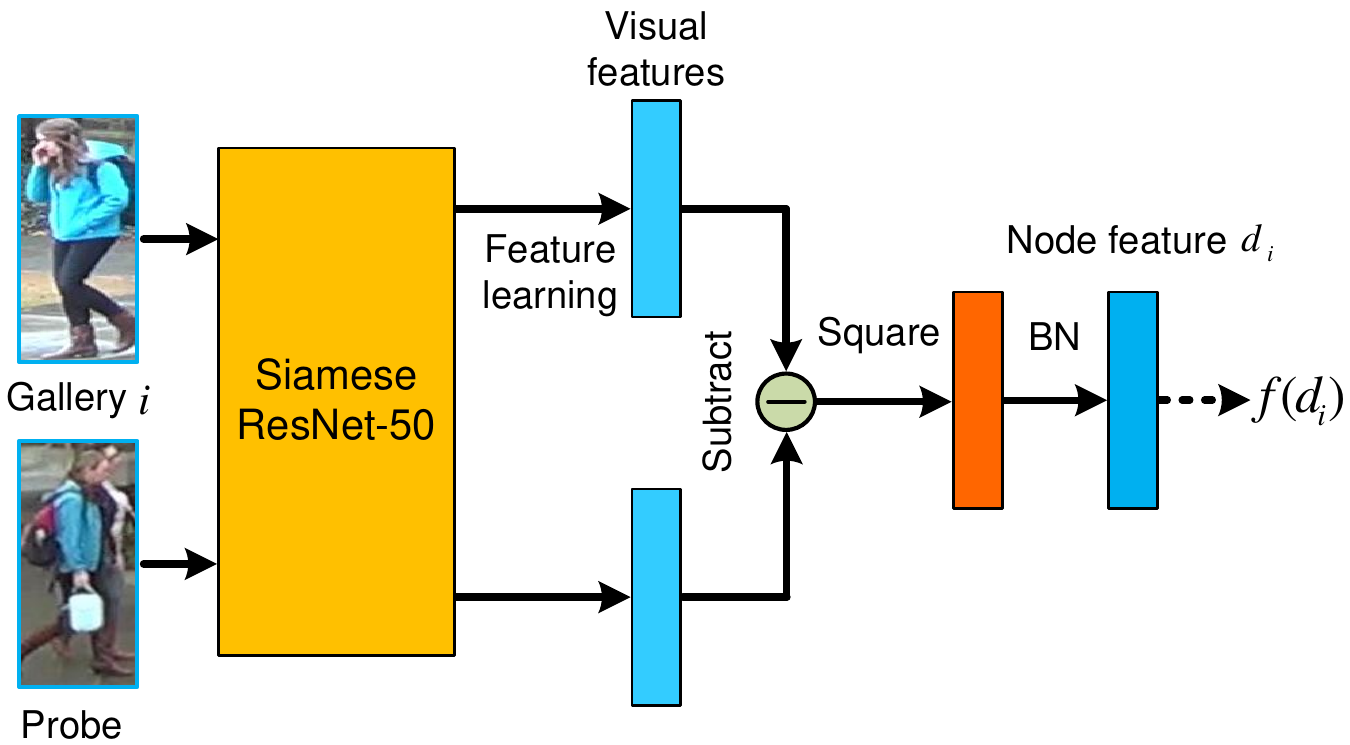} &\includegraphics[scale = 0.4]{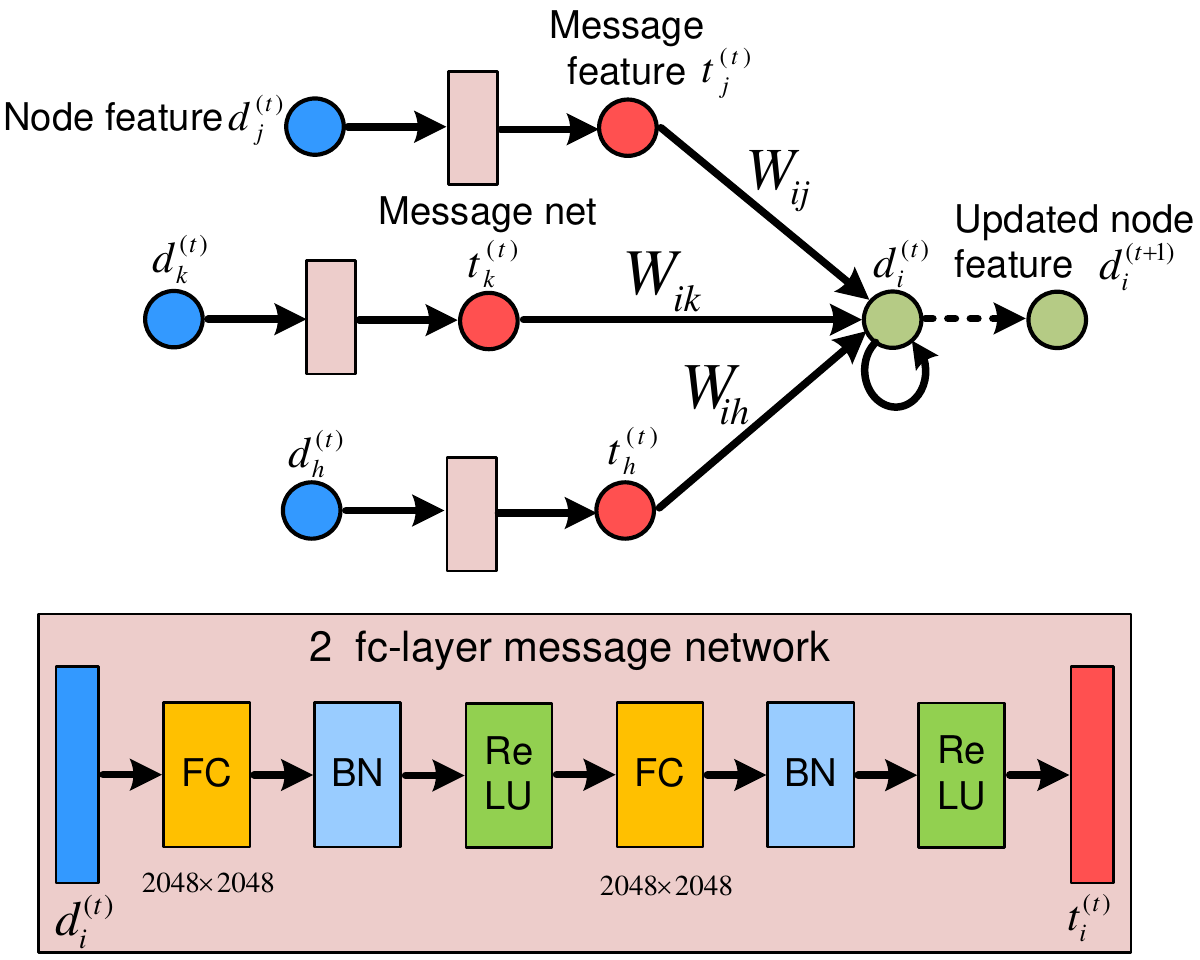}  \\
   & (a) Node input feature generating. & (b) Deep message passing of SGGNN.
\end{tabular}
\caption{The illustration of our base model and deep message passing of SGGNN. (a) Our base model is not only utilized for calculating the  probe-gallery pairs' similarity scores, but also for obtaining the gallery-gallery similarity scores, which could be utilized for deep message passing to update the relation features of probe-gallery pairs. (b) For passing more effective information, probe-gallery relation features $d_i$ are first fed into a 2 layer message network for feature encoding. With gallery-gallery similarity scores, the probe-gallery relation feature fusion could be deduced as a message passing and feature fusion schemes, which is defined as Eq.~\ref{eq:refine_walk}.}
\label{fig:embed}
\end{figure}
To enhance the initial pairwise relation features of a node with other nodes' information, we propose to propagate deeply learned messages between all connecting nodes. The node features are then updated as a weighted addition fusion of all input messages and the node's original features. The proposed relation feature fusion and updating is intuitive: using gallery-gallery similarity scores to guide the refinement of the probe-gallery relation features will make the relation features more discriminative and accurate, since the rich relation information among different pairs are involved. For instance, given one probe sample $p$ and two gallery samples $g_i$, $g_j$. Suppose that $(p, g_i)$ is a hard positive pair (node) while both $(p, g_j)$ and $(g_i, g_j)$ are relative easy positive pairs. Without any message passing among the nodes $(p, g_i)$ and $(p, g_j)$, the similarity score of $(p, g_i)$ is unlikely to be high. However, if we utilize the similarity of pair $(g_i, g_j)$ to guide the refinement of the relation features of the hard positive pair $(p, g_i)$, the refined features of $(p, g_i)$ will lead to a more proper similarity score. This relation feature fusion could be deduced as a message passing and feature fusion scheme. 

Before message passing begins, each node first encodes a deep message for sending to other nodes that are connected to it. The nodes' input relation features  $d_i$ are fed into a message network with  2 fully-connected layers with BN and  ReLU to generate deep message $t_i$, which is illustrated in Figure~\ref{fig:embed}(b). This process learns more suitable messages for node relation feature updating,

\begin{equation}\label{eq:transform}
   t_i = F(d_i) \quad \text{for }i=1,2,...,N,
\end{equation}
where $F$ denotes the 2 FC-layer subnetwork for learning deep messages for propagation.

After obtaining the edge weights $W_{ij}$ and deep message $t_i$ from each node, the updating scheme of node relation feature $d_i$ could be formulated as
\begin{equation}\label{eq:refine_walk}
   d_{i}^{(1)} = (1 -\alpha) d_{i}^{(0)} + \alpha \sum_{j = 1}^{N} W_{ij} t_{j}^{(0)} \quad \text{for} \ i=1,2,...,N,
\end{equation}
where $ d_{i}^{(1)}$ denotes the $i$-th refined relation feature, $d_{i}^{(0)}$  denotes the $i$-th input relation feature and $t_{j}^{(0)}$ denotes the deep message from node $j$. $\alpha$ represents the weighting parameter that balances fusion feature and original feature. 

Noted that such relation feature weighted fusion could be performed iteratively as follows,
\begin{equation}\label{eq:refine_walk_iter}
   d_{i}^{(t)} =(1 - \alpha) d_{i}^{(t-1)} + \alpha \sum_{j = 1}^{N} W_{ij} t_{j}^{(t-1)} \quad \text{for} \  i=1,2,...,N,
\end{equation}
where $t$ is the iteration number. The refined relation feature $d_i^{(t)}$ could substitute then relation feature $d_i$ in Eq.~(\ref{eq:loss}) for loss computation and training the SGGNN. For training,  Eq.~(\ref{eq:refine_walk_iter}) can be unrolled via back propagation through structure.

In practice, we found that the performance gap between iterative feature updating of multiple iterations and updating for one iteration is negligible. So we adopt Eq. ~(\ref{eq:refine_walk}) as our relation feature fusion in both training and testing stages. After relation feature updating, we feed the relation features of probe-gallery image pairs to a linear classifier with sigmoid function for obtaining the similarity score and trained with the same binary cross-entropy loss (Eq.~(\ref{eq:loss})).

\subsection{Relations to Conventional GNN}
\label{sec:advan}
In our proposed SGGNN model, the similarities among gallery images are served as fusion weights on the graph for nodes' feature fusion and updating. These similarities are vital for refining the probe-gallery relation features. In conventional GNN~\cite{yan2018spatial,niepert2016learning} models, the feature fusion weights are usually modeled as a nonlinear function $h(d_i, d_j)$ that measures compatibility between two nodes $d_i$ and $d_j$. The feature updating will be  
\begin{equation}\label{eq:gnn}
   d_{i}^{(t)} =(1 - \alpha) d_{i}^{(t-1)} + \alpha \sum_{j = 1}^{N} h(d_i, d_j) t_{j}^{(t-1)} \quad \text{for} \  i=1,2,...,N.
\end{equation}
They lack directly label supervision and are only indirectly learned via back-propagation errors. However, in our case, such a strategy does not fully utilize the similarity ground-truth between gallery images. To overcome such limitation, we propose to use similarity scores $S(g_i, g_j)$ between gallery images $g_i$ and $g_j$ with directly training label supervision to serve as the node feature fusion weights in Eq.~(\ref{eq:refine_walk}). Compared with conventional setting of GNN Eq.~(\ref{eq:gnn}), these direct and rich supervisions of gallery-gallery similarity could provide feature fusion with more accurate information.

\subsection{Implementation Details} 
\label{sec:imp}

Our proposed SGGNN is based on ResNet-50~\cite{he2016deep} pretrained on ImageNet~\cite{deng2009imagenet}. The input images are all resized to $256 \times 128$. Random flipping and random erasing~\cite{zhong2017random} are utilized for data augmentation. We will first pretrain the base Siamese CNN model, we adopt an initial learning rate of 0.01 on all three datasets and reduce the learning rate by 10 times after 50 epochs. The learning rate is then fixed for another 50 training epochs. The weights of linear classifier for obtaining the gallery-gallery similarities  is initialized with the weights of linear classifier we trained in the base model pretraining stage. To construct each mini-batch as a combination of a probe set and a gallery set, we randomly sample images according to their identities. First we randomly choose $M$ identities in each mini-batch. For each identity, we randomly choose $K$ images belonging to this identity. Among these $K$ images of one person, we randomly choose one of them as the probe image and leave the rest of them as gallery images. As a result, a $K \times M$ sized mini-batch consists of a size $K$ probe set and a size $K \times (M-1)$ gallery set. In the training stage, $K$ is set to 4 and $M$ is set to 48, which results in a mini-batch size of 192. In the testing stage, for each probe image, we first utilize $l2$ distance between probe image feature and gallery image features by the trained ResNet-50 in our SGGNN to obtain the top-100 gallery images, then we use SGGNN for obtaining the final similarity scores. We will go though all the identities in each training epoch and Adam algorithm~\cite{kingma2014adam} is utilized for optimization. 

We then finetune the overall SGGNN model end-to-end, the input node features for overall model are the subtracted features of base model. Note that for gallery-gallery similarity estimation $S(g_i, g_j)$, the rich labels of gallery images are also used as training supervision. we train the overall network with a learning rate of $10^{-4}$ for another 50 epochs and the balancing weight $\alpha$ is set to 0.9.

\section{Experiments}
\subsection{Datasets and Evaluation Metrics}
To validate the effectiveness of our proposed approach for person re-identification. The experiments and ablation study are conducted on three large public datasets.

\textbf{CUHK03}~\cite{li2014deepreid} is a person re-identification dataset, which contains 14,097 images of 1,467 person captured by two cameras from the campus. We utilize its manually annotated images in this work.

\textbf{Market-1501}~\cite{zheng2015scalable} is a large-scale dataset, which contains multi-view person images for each identity. It consists of 12,936 images for training and 19,732 images for testing. The test set is divided into a gallery set that contains 16,483 images and a probe set that contains 3,249 images. There are totally 1501 identities in this dataset and all the person images are obtained by DPM detector~\cite{felzenszwalb2010object}.

\textbf{DukeMTMC}~\cite{ristani2016MTMC} is collected from campus with 8 cameras, it originally contains more than  2,000,000 manually annotated frames. There are some extensions for DukeMTMC dataset for person re-identification task. In this paper, we follow the setting of~\cite{zheng2017unlabeled}. It utilizes 1404 identities, which appear in more than two cameras. The training set consists of 16,522 images with 702 identities and test set contains 19,989 images with 702 identities. 

We adopt mean average precision (mAP) and CMC top-1, top-5, and top-10 accuracies as evaluation metrics. For each dataset, we just adopt the original evaluation protocol that the dataset provides. In the experiments, the query type is single query.
\begin{table}
\setlength{\tabcolsep}{4pt}
   \small
   \begin{center}
   \caption{mAP, top-1, top-5, and top-10 accuracies by compared methods on the CUHK03 dataset~\cite{li2014deepreid}.}
   \label{tab:cuhk}
      \begin{tabular}{lccccc}
         \toprule
         \multirow{2}{*}{Methods}&
         \multirow{2}{*}{Conference}&
         \multicolumn{4}{c}{CUHK03~\cite{li2014deepreid}}\\
         & &mAP&top-1&top-5&top-10\\
         \midrule
         Quadruplet Loss~\cite{Chen_2017_CVPR}&CVPR 2017& -&75.5&95.2&99.2\\
         OIM Loss~\cite{xiao2017joint} &CVPR 2017& 72.5 & 77.5  & - & - \\
         SpindleNet~\cite{zhao2017spindle} &CVPR 2017& -&88.5&97.8 & 98.6\\
         MSCAN~\cite{Li_2017_CVPR} &CVPR 2017& -&74.2&94.3&97.5\\
         SSM~\cite{bai2017scalable}&CVPR 2017& -&76.6&94.6 & 98.0\\
         k-reciprocal~\cite{Zhong_2017_CVPR}&CVPR 2017&67.6 &61.6& - & - \\
         VI+LSRO~\cite{zheng2017unlabeled}&ICCV 2017&87.4&84.6&97.6&98.9\\
         SVDNet~\cite{Sun_2017_ICCV} &ICCV 2017&84.8& 81.8&95.2&97.2\\
         OL-MANS~\cite{Zhou_2017_ICCV}&ICCV 2017& -&61.7&88.4&95.2\\
         Pose Driven~\cite{Su_2017_ICCV}&ICCV 2017& -&88.7&98.6&\textbf{99.6}\\
         Part Aligned~\cite{zhao2017deeply}&ICCV 2017& -& 85.4 &97.6&99.4\\
         HydraPlus-Net~\cite{Liuxihui_2017_ICCV}&ICCV 2017& - & 91.8 &98.4&99.1\\
         MuDeep~\cite{qian2017multi}&ICCV 2017& -&76.3 &96.0 &98.4\\
         JLML~\cite{gong2017person}&IJCAI 2017& -&83.2&98.0 & 99.4\\
         MC-PPMN~\cite{mao2018multi}&AAAI 2018& -&86.4&98.5 & \textbf{99.6}\\
         Proposed SGGNN & ~&\textbf{94.3} &\textbf{95.3} & \textbf{99.1} & \textbf{99.6}\\
         \bottomrule
      \end{tabular}
   \end{center}
\end{table}

\begin{table}
   \small
   \setlength{\tabcolsep}{3pt}
      \begin{center}
      \caption{mAP, top-1, top-5, and top-10 accuracies of compared methods on the Market-1501 dataset~\cite{zheng2015scalable}.}
      \label{tab:market}
      \begin{tabular}{lccccc}
         \toprule
         \multirow{2}{*}{Methods}&
         \multirow{2}{*}{Reference}&
         \multicolumn{4}{c}{Market-1501~\cite{zheng2015scalable}}\\
         & &mAP&top-1&top-5&top-10\\
         \midrule
         OIM Loss~\cite{xiao2017joint}&CVPR 2017& 60.9 & 82.1  & - & - \\
         SpindleNet~\cite{zhao2017spindle}&CVPR 2017& -&76.9&91.5&94.6\\
         MSCAN~\cite{Li_2017_CVPR} &CVPR 2017&53.1&76.3& - & - \\
         SSM~\cite{bai2017scalable}&CVPR 2017&68.8&82.2& - & -\\
         k-reciprocal~\cite{Zhong_2017_CVPR}&CVPR 2017& 63.6&77.1& - & - \\
         Point 2 Set~\cite{Zhou_2017_CVPR} &CVPR 2017&44.3&70.7& - & - \\
         CADL~\cite{Lin_2017_CVPR} & CVPR 2017&47.1&73.8& - & - \\
         VI+LSRO~\cite{zheng2017unlabeled} &ICCV 2017&66.1&84.0& -& -\\
         SVDNet~\cite{Sun_2017_ICCV}&ICCV 2017& 62.1&82.3&92.3&95.2\\
         OL-MANS~\cite{Zhou_2017_ICCV} &ICCV 2017& -&60.7& -& -\\
         Pose Driven~\cite{Su_2017_ICCV}&ICCV 2017&63.4&84.1&92.7&94.9\\
         Part Aligned~\cite{zhao2017deeply}&ICCV 2017& 63.4 & 81.0 &92.0&94.7\\
         HydraPlus-Net~\cite{Liuxihui_2017_ICCV}&ICCV 2017& - & 76.9 &91.3&94.5\\
         JLML~\cite{gong2017person}&IJCAI 2017&65.5&85.1& - & -\\
		 HA-CNN~\cite{li2018harmonious}&CVPR 2018&75.7&91.2& -& ~-\\
         Proposed SGGNN &~& \textbf{82.8} &\textbf{92.3} &\textbf{96.1} & \textbf{97.4}\\
         \bottomrule
      \end{tabular}
   \end{center}
\end{table}

\subsection{Comparison with State-of-the-art Methods}

\subsubsection{Results on CUHK03 dataset.}
The results of our proposed method and other state-of-the-art methods are represented in Table~\ref{tab:cuhk}. The mAP and top-1 accuracy of our proposed method are 94.3\% and 95.3\%, respectively. Our proposed method outperforms all the compared methods.

Quadruplet Loss~\cite{Chen_2017_CVPR} is modified based on triplet loss. It aims at obtaining correct orders for input pairs and pushing away negative pairs from positive pairs. Our proposed method outperforms quadruplet loss 19.8\% in terms of top-1 accuracy. OIM Loss~\cite{xiao2017joint} maintains a look-up table. It compares distances between mini-batch samples and all the entries in the table. to learn features of person image. Our approach improves OIM Loss by 21.8\% and 17.8\% in terms of mAP and CMC top-1 accuracy. SpindleNet~\cite{zhao2017spindle} considers body structure information for person re-identification. It incorporates body region features and features from different semantic levels for person re-identification. Compared with SpindleNet, our proposed method increases 6.8\% for top-1 accuracy. MSCAN~\cite{li2017learning} stands for Multi-Scale ContextAware Network. It adopts multiple convolution kernels with different receptive fields to obtain multiple feature maps. The dilated convolution is utilized for decreasing the correlations among convolution kernels. Our proposed method gains 21.1\% in terms of top-1 accuracy. SSM stands for Smoothed Supervised Manifold~\cite{bai2017scalable}. This approach tries to obtain the underlying manifold structure by estimating the similarity between two images in the context of other pairs of images in the post-processing stage, while the proposed SGGNN utilizes instance relation information in both training and testing stages. SGGNN outperforms SSM approach by 18.7\% in terms of top-1 accuracy. k-reciprocal ~\cite{Zhong_2017_CVPR} utilized gallery-gallery similarities in the testing stage and uses a smoothed Jaccard distance for refining the ranking results. In contrast, SGGNN exploits the gallery-gallery information in the training stage for feature learning. As a result, SGGNN gains 26.7\% and 33.7\% increase in terms of mAP and top-1 accuracy. 

\subsubsection{Results on Market-1501 dataset.}
On Market-1501 dataset, our proposed methods outperforms significantly state-of-the-art methods. SGGNN achieves mAP of 82.8\% and top-1 accuracy of 92.3\% on Market-1501 dataset. The results are shown in Table~\ref{tab:market}.

HydraPlus-Net~\cite{Liuxihui_2017_ICCV} is proposed for better exploiting the global and local contents with multi-level feature fusion of a person image. Our proposed method outperforms HydraPlus-Net by 15.4 for top-1 accuracy. JLML~\cite{gong2017person} stands for Joint Learning of Multi-Loss. JLML learns both global and local discriminative features in different context and exploits complementary advantages jointly. Compared with JLML, our proposed method gains 17.3 and 7.2 in terms of mAP and top-1 accuracy. HA-CNN~\cite{li2018harmonious} attempts to learn hard region-level and soft pixel-level attention simultaneously with arbitrary person bounding boxes and person image features. The proposed SGGNN outperforms HA-CNN by 7.1\% and 1.1\% with respect to mAP and top-1 accuracy.

\subsubsection{Results on DukeMTMC dataset.}
In Table~\ref{tab:duke}, we illustrate the performance of our proposed SGGNN and other state-of-the-art methods on DukeMTMC~\cite{ristani2016MTMC}. Our method outperforms all compared approaches. Besides approaches such as OIM Loss and SVDNet, which have been introduced previously, our method also outperforms Basel+LSRO, which integrates GAN generated data and ACRN that incorporates person of attributes for person re-identification significantly. These results illustrate the effectiveness of our proposed approach. 
\begin{table}
   \small
   \setlength{\tabcolsep}{4pt}
   \begin{center}
     \caption{mAP, top-1, top-5, and top-10 accuracies by compared methods on the DukeMTMC dataset~\cite{ristani2016MTMC}.}
     \label{tab:duke}
      \begin{tabular}{lccccc}
         \toprule
         \multirow{2}{*}{Methods}&
         \multirow{2}{*}{Reference}&
         \multicolumn{4}{c}{DukeMTMC~\cite{ristani2016MTMC}}\\
         & &mAP&top-1&top-5&top-10\\
         \midrule
         BoW+KISSME~\cite{zheng2015scalable} &ICCV 2015& 12.2 &25.1& -& -\\
         LOMO+XQDA~\cite{liao2015person} &CVPR 2015& 17.0 &30.8& -& -\\
         ACRN~\cite{schumann2017person}&CVPRW 2017& 52.0 &72.6&84.8&88.9\\
         OIM Loss~\cite{xiao2017joint}&CVPR 2017& 47.4 & 68.1  & - & - \\
         Basel.+LSRO~\cite{zheng2017unlabeled} &ICCV 2017&47.1&67.7& -& -\\
         SVDNet~\cite{Sun_2017_ICCV}&ICCV 2017&56.8 &76.7&86.4&89.9\\
         Proposed SGGNN & ~&\textbf{68.2} &\textbf{81.1} &\textbf{88.4} & \textbf{91.2}\\
         \bottomrule
      \end{tabular}
   \end{center}
\end{table}

\subsection{Ablation Study}
To further investigate the validity of SGGNN, we also conduct a series of ablation studies on all three datasets. Results are shown in Table~\ref{tab:abl}.

We treat the siamese CNN model that directly estimates pairwise similarities from initial node features introduced in Section~\ref{sec:pro_for} as the base model. We utilize the same base model and compare with other approaches that also take inter-gallery image relations in the testing stage for comparison. We conduct k-reciprocal re-ranking~\cite{Zhong_2017_CVPR} with the image visual features learned by our base model. Compared with SGGNN approach, The mAP of k-reciprocal approach drops by 4.3\%, 4.4\%, 3.5\% for Market-1501, CUHK03, and DukeMTMC datasets. The top-1 accuracy also drops by 0.8\%, 3.1\%, 1.2\% respectively. Except for the visual features, base model could also provides us raw similarity scores of probe-gallery pairs and gallery-gallery pairs. A random walk~\cite{bai2017scalable} operation could be conducted to refine the probe-gallery similarity scores with gallery-gallery similarity scores with a closed-form equation. Compared with our method, The performance of random walk drops by 3.6\%, 4.1\%, and 2.2\% in terms of mAP, 0.8\%, 3.0\%, and 0.8\% in terms of top-1 accuracy. Such results illustrate the effectiveness of end-to-end training with deeply learned message passing within SGGNN. We also validate the importance of learning  visual feature fusion weight with gallery-gallery similarities guidance. In Section~\ref{sec:advan}, we have introduced that in the conventional GNN, the compatibility between two nodes $d_i$ and $d_j$, $h(d_i, d_j)$ is calculated by a non-linear function, inner product function without direct gallery-gallery supervision. We therefore remove the directly gallery-gallery supervisions and train the model with weight fusion approach in Eq.~(\ref{eq:gnn}) , denoted by \textit{Base Model + SGGNN w/o SG}.  The performance drops by 1.6\%, 1.6\%, and 0.9\% in terms of mAP. The top-1 accuracies drops 1.7\%, 2.6\%, and 0.6\% compared with our SGGNN approach, which illustrates the importance of involving rich gallery-gallery labels in the training stage. 

To demonstrate that our proposed model SGGNN also learns better visual features by considering all probe-gallery relations, we evaluate the re-identification performance by directly calculating the $l_2$ distance between different images' visual feature vectors outputted by our trained ResNet-50 model on three datasets. The results by visual features learned with base model and the conventional GNN approach are illustrated in Table~\ref{tab:feature}. Visual features by our proposed SGGNN outperforms the compared base model and conventional GNN setting significantly, which demonstrates that SGGNN also learns more discriminative and robust features.

\subsection{Sensitivity Analysis}
We tried training our SGGNN with different $K$ and also testing with different top-$K$ choices (Table \ref{table:sense}, rows 2-5). Results show that higher top-$K$ slightly increases accuracy but also increases computational cost.

\begin{table}
\small
   \begin{center}
   \caption{Ablation studies on the Market-1501~\cite{zheng2015scalable}, CUHK03~\cite{li2014deepreid} and DukeMTMC~\cite{ristani2016MTMC} datasets.}
   \label{tab:abl}
      \begin{tabular}{lccccccc}
         \toprule
         \multirow{2}{*}{Methods}&
         \multicolumn{2}{c}{Market-1501~\cite{zheng2015scalable}}&
         \multicolumn{2}{c}{CUHK03~\cite{li2014deepreid}} &
         \multicolumn{2}{c}{DukeMTMC~\cite{ristani2016MTMC}}\\
         &mAP&top-1&mAP&top-1&mAP&top-1\\
         \midrule
         Base Model &76.4 &91.2 &88.9&91.1 &61.8&78.8\\
         Base Model + k-reciprocal ~\cite{Zhong_2017_CVPR}&78.5&91.5&89.9&92.2&64.7&79.9\\
         Base Model + random walk~\cite{bai2017scalable}&79.2 &91.5 & 90.2 & 92.3& 66.0 & 80.3\\
         Base Model + SGGNN w/o SG &81.2&90.6&92.7&93.6&67.3&80.5\\
         Base Model + SGGNN & \textbf{82.8} &\textbf{92.3} & \textbf{94.3} & \textbf{95.3} &\textbf{68.2}&\textbf{81.1}\\
         \bottomrule
      \end{tabular}
   \end{center}
\end{table}

\begin{table}
\small
   \begin{center}
   \caption{Performances of estimating probe-gallery similarities by $l_2$ feature distance on the Market-1501~\cite{zheng2015scalable}, CUHK03~\cite{li2014deepreid} and DukeMTMC~\cite{ristani2016MTMC} datasets.}
   \label{tab:feature}
      \begin{tabular}{lccccccc}
         \toprule
         \multirow{2}{*}{Model}&
         \multicolumn{2}{c}{Market-1501~\cite{zheng2015scalable}}&
         \multicolumn{2}{c}{CUHK03~\cite{li2014deepreid}} &
         \multicolumn{2}{c}{DukeMTMC~\cite{ristani2016MTMC}}\\
         &mAP&top-1&mAP&top-1&mAP&top-1\\
         \midrule
         Base Model &74.6 &90.4 &87.6&91.0 &60.3&77.6\\
         Base Model + SGGNN w/o SG &75.4 &90.4&87.7&91.5&61.7&78.1\\
         Base Model + SGGNN & \textbf{76.7} &\textbf{91.5} & \textbf{88.1} & \textbf{93.6} &\textbf{64.6}&\textbf{79.1}\\
         \bottomrule
      \end{tabular}
   \end{center}
\end{table}

\begin{table}
\begin{center}
\caption{Performances of different $K$ and top-$K$ choices.}
\label{table:sense}
\begin{tabular}{cccc|cccccc}
         \toprule
         \multicolumn{4}{c}{Parameters Settings}&
         \multicolumn{2}{c}{Market-1501}&
         \multicolumn{2}{c}{CUHK03} &
         \multicolumn{2}{c}{DukeMTMC}\\
         Top-$K$ & $K$ &  $\alpha$ &  $t$ &mAP&top-1&mAP&top-1&mAP&top-1\\
         \midrule
         Top-$100$ & $4$ &  0.9 & 1 & 82.8 &92.3 &94.3 & 95.3 &68.2&81.1\\
         \hline
         Top-100 & $3$ &  0.9 & 1 &82.0 &91.7&94.1&95.2&68.2&80.8\\
         Top-100 &$5$ &  0.9 & 1 &82.1 &91.8&94.2&95.2&68.0&80.6\\
         Top-50 &$4$ &  0.9 & 1 &80.7 &91.3&93.7&95.1&66.6&79.8\\
         Top-150 &$4$ &  0.9 & 1 &83.6 &92.0&94.5&95.3&71.8&83.5\\
         \hline
         Top-100 & $4$ & 0.9 & 2 &82.9 &91.3 & 95.1&96.1 &68.9 &81.7 \\
         Top-100 & $4$ & 0.9 & 3 &81.3 &89.3 &95.4 &96.0 &69.0 &81.9 \\
         \hline
         Top-100 & $4$ & 0.5 & 1 &79.8 &91.4&92.4&94.2&66.6&81.0\\
         Top-100 & $4$ & 0.95 & 1 &82.8 &92.8&94.3&95.4&68.3&81.6\\
         \bottomrule
\end{tabular}

\end{center}
\end{table}

\section{Conclusion}
In this paper, we propose Similarity-Guided Graph Neural Neural to incorporate the rich gallery-gallery similarity information into training process of person re-identification. Compared with our method, most previous attempts conduct the updating of probe-gallery similarity in the post-process stage, which could not benefit the learning of visual features. For conventional Graph Neural Network setting, the rich gallery-gallery similarity labels are ignored while our approach utilized all valuable labels to ensure the weighted deep message fusion is more effective. The overall performance of our approach and ablation study illustrate the effectiveness of our proposed method.

\section{Acknowledgements}
This work is supported by SenseTime Group Limited, the General Research Fund sponsored by the Research Grants Council of Hong Kong (Nos. CUHK14213616, CUHK14206114, CUHK14205615, CUHK14203015, CUHK14239816, CUHK419412, CUHK14207814, CUHK14208417, CUHK14202217), the Hong Kong Innovation and Technology Support Program (No. ITS/121/15FX).

\clearpage

{\small
\bibliographystyle{ieee.bst}
\bibliography{egbib.bib}
}

\end{document}